\documentclass{article} % For LaTeX2e
\pdfoutput=1

\PassOptionsToPackage{numbers}{natbib}

\usepackage[final]{neurips_2018}
\usepackage{hyperref}       % hyperlinks
\usepackage{url}            % simple URL typesetting

\usepackage{amsmath}
\usepackage{amsthm}
\usepackage{amssymb}
\usepackage{units}
\usepackage{graphicx}
\usepackage{tabularx}
\usepackage{caption}
\usepackage{verbatim}
\usepackage{hhline}
\usepackage{xcolor}
\usepackage{soul}

\title{Long short-term memory and learning-to-learn in networks of spiking neurons}

\author{Guillaume Bellec*, Darjan Salaj*, Anand Subramoney*, Robert Legenstein \& Wolfgang Maass \\
Institute for Theoretical Computer Science\\
Graz University of Technology, Austria \\
\texttt{\{bellec,salaj,subramoney,legenstein,maass\}@igi.tugraz.at} \\
\text{* equal contributions}
}

\begin{document}

\maketitle

\begin{abstract}

Recurrent networks of spiking neurons (RSNNs) underlie the astounding computing and learning capabilities of the brain. But computing and learning capabilities of RSNN models have remained poor, at least in comparison with artificial neural networks (ANNs). We address two possible reasons for that. One is that RSNNs in the brain are not randomly connected or designed according to simple rules, and they do not start learning as a tabula rasa network. Rather, RSNNs in the brain were optimized for their tasks through evolution, development, and prior experience. Details of these optimization processes are largely unknown. But their functional contribution can be approximated through powerful optimization methods, such as backpropagation through time (BPTT).

A second major mismatch between RSNNs in the brain and models is that the latter only show a small fraction of the dynamics of neurons and synapses in the brain. We include neurons in our RSNN model that reproduce one prominent dynamical process of biological neurons that takes place at the behaviourally relevant time scale of seconds: neuronal adaptation. We denote these networks as LSNNs because of their Long short-term memory. The inclusion of adapting neurons drastically increases the computing and learning capability of RSNNs if they are trained and configured by deep learning (BPTT combined with a rewiring algorithm that optimizes the network architecture). In fact, the computational performance of these RSNNs approaches for the first time that of LSTM networks. In addition RSNNs with adapting neurons can acquire abstract knowledge from prior learning in a Learning-to-Learn (L2L) scheme, and transfer that knowledge in order to learn new but related tasks from very few examples. We demonstrate this for supervised learning and reinforcement learning.

%Networks of spiking neurons (SNNs) are frequently studied as models for networks of
%neurons in the brain, but also as paradigm for novel energy efficient computing hardware. In principle they are especially suitable for computations in the temporal domain, such as speech processing, because their computations are carried out via events in time and space. But so far they have been lacking the capability to preserve information for longer time spans during a computation, like a register of a digital computer. This function is provided to artificial neural networks through Long Short-Term Memory (LSTM) units. We show here that SNNs attain similar capabilities if one includes adapting neurons in the network. Adaptation denotes an increase of the firing threshold of a neuron after preceding firing. A substantial fraction of neurons in the neocortex of rodents and humans has been found to be adapting. It turns out that if adapting neurons are integrated in a suitable
%manner into the architecture of SNNs, the performance of these enhanced SNNs, which we call LSNNs, for computation in the temporal domain approaches that of artificial neural networks with LSTM-units. In addition, the computing and learning capabilities of LSNNs can be substantially enhanced through learning-to-learn (L2L) methods from machine learning, that have so far been applied primarily to LSTM networks and apparently never to SSNs.
\end{abstract}

\section{Introduction}\label{sec:model}

Recurrent networks of spiking neurons (RSNNs) are frequently studied as
models for networks of neurons in the brain. In principle, they should be
especially well-suited for computations in the temporal domain, such as
speech processing, as their computations are carried out via spikes, i.e.,
events in time and space. 
But the performance of RSNN models has remained suboptimal also for temporal processing tasks.
One difference between RSNNs in the brain and RSNN models is that RSNNs in the brain have been optimized for their function through long evolutionary processes, complemented by a sophisticated learning curriculum during development. Since most details of these biological processes are currently still unknown, we asked whether deep learning is able to mimic these complex optimization processes on a functional level for RSNN models.
We used BPTT as the deep learning method for network optimization. Backpropagation has been adapted previously for feed forward networks with binary activations in \cite{courbariaux_binarized_2016,esser_convolutional_2016}, and we adapted BPTT to work in a similar manner for RSNNs. In order to also optimize the connectivity of RSNNs, we augmented BPTT with DEEP R, a biologically inspired heuristic for synaptic rewiring \cite{kappel2017reward,deepr}.  
Compared to LSTM networks, RSNNs tend to have inferior short-term memory capabilities. Since neurons in the brain are equipped with a host of dynamics processes on time scales larger than a few dozen ms \cite{hasson2015hierarchical}, we enriched the inherent 
%time constants through simple models for adapting spiking neurons.
dynamics of neurons in our model by a standard neural adaptation process.

We first show (section \ref{sec:comp_cap}) that this approach produces new computational performance levels of RSNNs for two common benchmark tasks: Sequential MNIST and TIMIT (a speech processing task). 
We then show that it makes L2L applicable to RSNNs (section \ref{sec:L2L}), similarly as for LSTM networks. 
In particular, we show that meta-RL \cite{wang2016learning,duan2016rl} produces new motor control capabilities of RSNNs (section \ref{sec:meta-RL}). This result links a recent abstract model for reward-based learning in the brain \cite{WangETAL:18} to spiking activity. In addition, we show that RSNNs with sparse connectivity and sparse firing activity of 10-20 Hz (see Fig. 1D, 2D, S1C) can solve these and other tasks. Hence these RSNNs compute with spikes, rather than firing rates.%biologically more realistic models on the neural network level.

%Our results show that L2L induces a powerful new method for fast learning in RSNNs. L2L results for RSNNs have the advantage that many features and fingerprints of resulting new learning algorithms for the network, such as neural activity (spike trains), changes in the excitability of neurons and synaptic weights, and changes in the network dynamics through learning can be compared with measurements from networks of neurons in the brain.

%At the same time, the resulting new methods for learning nonlinear functions from few examples are 
The superior computing and learning capabilities of LSNNs suggest that they are also of interest for implementation in spike-based neuromorphic chips such as Brainscales 
\cite{schemmel2010wafer}, SpiNNaker \cite{furber2013overview}, True North \cite{esser_convolutional_2016}, chips from
ETH Z\"urich \cite{qiao2015reconfigurable}, and Loihi \cite{davies2018loihi}. In particular, nonlocal learning rules such as
backprop are challenges for some of these neuromorphic devices (and for many brain models).
Hence alternative methods for RSNN learning of nonlinear functions are needed. We show in sections \ref{sec:L2L} and \ref{sec:meta-RL} that L2L can be used to generate RSNNs that learn very efficiently even in the absence of synaptic plasticity.

%\todo{Added from rebuttal}
\textbf{Relation to prior work:} We refer to \cite{eliasmith2013build,depasquale2016using,huh2017gradient,nicola2017supervised} for summaries of preceding results on computational capabilities of RSNNs. %and more detailed summaries of the state of the art.
The focus there was typically on the generation of dynamic patterns.
%insert Wm 8.10.18
Such tasks are not addressed in this article, but it will be shown in \cite{bellec2018learncompute} that LSNNs provide an alternative model to \cite{nicola2017supervised} for the generation of complex temporal patterns.
Huh et al.~\cite{huh2017gradient} applied gradient descent to recurrent networks of spiking neurons. There, neurons without a leak were used. Hence, the voltage of a neuron could used in that approach to store information over an unlimited length of time.

We are not aware of previous attempts to bring the performance of RSNNs for time series classification into the performance range of LSTM networks. We are also not aware of any previous literature on applications of L2L to SNNs.

\section{LSNN model}\label{sec:LSNNs}

%\textbf{The LSNN model: }

% ask wolfgang
%The considered RSNNs consist of standard leaky integrate-and-fire (LIF) neuron models (see Supplementary Information). But since RSNNs in the brain are equipped with a host of dynamics processes on larger time scales than a few dozen ms, 
%in the LSNN consists of standard neurons of this type, whereas 
%Einf2 - WM 15.5.18
Neurons and synapses in common RSNN models are missing many of the dynamic processes found in their biological counterparts, especially those on larger time scales. We integrate one of them into our RSNN model: neuronal adaptation. It is well known that a substantial fraction of excitatory neurons in the brain are adapting, with diverse time constants, see e.g. the Allen Brain Atlas for data from the neocortex of mouse and humans. We refer to the resulting type of RSNNs as Long short-term memory Spiking Neural Networks (LSNNs).
LSNNs consist of a population $R$ of integrate-and-fire (LIF) neurons (excitatory and inhibitory), and a second population $A$ of LIF excitatory neurons whose excitability is temporarily reduced through preceding firing activity, i.e., these neurons are adapting (see Fig.~\ref{fig:mnist}C and Suppl.). %It is well known that a substantial fraction of excitatory neurons in the brain is adapting, with diverse time constants, see e.g. the Allen Brain Atlas \cite{allen-featuresearch} for data from the neocortex of mouse and humans.
Both populations $R$ and $A$ receive spike trains from a population $X$ of external input neurons. Results of computations are read out by a population $Y$ of external linear readout neurons, see Fig.~\ref{fig:mnist}C.

%In neuroscience the term neural adaptation refers to a broad range of phenomena and mechanisms. It was already discovered
%in the 19th century that the intensity of a constant stimulus tends to cause a diminishing sensation, see e.g. the
%Wikipedia article ``Neural adaptation''. A closer look shows that these phenomena take place simultaneously on several time scales, from hundreds of ms to s, hours, or even days. In the context of neuron models, adaptation refers to changes in the firing threshold or subthreshold currents of a neuron that reduce its excitability in response to its own firing \cite{gerstner2014neuronal}. Experimental data on that are documented for example in the form of a distribution of measured adaptation indices of neurons for large sets of neurons, both from mouse and humans, in the online Allen Brain Atlas \cite{allen-featuresearch}. This adaptation index measures the increase of interspike intervals of a neuron that receives a constant input current. 
Common ways for fitting models for adapting neurons to data are described in \cite{gerstner2014neuronal,pozzorini2015automated,gouwens2018systematic,teeter2018generalized}. We are using here the arguably simplest model: We assume that the firing threshold $B_{j}(t)$ of neuron $j$ increases by some fixed amount $\beta / \tau_{a,j}$ for each spike of this neuron $j$, and then decays exponentially back to a baseline value $b_{j}^0$ with a time constant $\tau_{a,j}$. Thus the threshold dynamics for a discrete time step of $\delta t = 1$ ms reads as follows
\begin{eqnarray}
B_{j}(t) 		& = & b_{j}^0 + \beta b_j(t) , \\
b_{j}(t + \delta t) 	& = & \rho_j b_{j}(t) + (1 - \rho_j) z_j(t) ,
\end{eqnarray}
where $\rho_j=\exp(-\frac{\delta t}{\tau_{a,j}})$ and $z_j(t)$ is the spike train of neuron $j$ assuming values in $\{0,\frac{1}{\delta t}\}$.
Note that this dynamics of thresholds of adaptive spiking neurons is similar to the dynamics of the state of context neurons in \cite{mikolov2014learning}.
It generally suffices to place the time constant of adapting neurons into the desired range for short-term memory (see Suppl. for specific values used in each experiment).
%Einf3 - WM 15.5.18
%We refer to the resulting type of RSNNs as Long short-term memory Spiking Neural Networks (LSNNs).

%\textbf{Deep learning and L2L for LSNNs:}\label{sec:DL}
% \section{BPTT together with DEEP R trains and configures RSNNs and LSNNs}\label{sec:BPTT}
\section{Applying BPTT with DEEP R to RSNNs and LSNNs}\label{sec:BPTT}

%Insert 8 WM 21.3.18
We optimize the synaptic weights, and in some cases also the connectivity matrix of an LSNN for specific ranges of tasks. 
%Einfügung 11 WM 23.3.18
The optimization algorithm that we use, backpropagation through time (BPTT), is not claimed to be biologically realistic. But like evolutionary and developmental processes, BPTT can optimize LSNNs for specific task ranges.
%For the three experiments that we report here we used backprop through time (BPTT). 
Backpropagation (BP) had already been
applied in \cite{courbariaux_binarized_2016} and \cite{esser_convolutional_2016} to feedforward networks of spiking
neurons. In these approaches, the gradient is backpropagated
through spikes by replacing the non-existent derivative of the membrane potential at the time of a
spike by a pseudo-derivative that smoothly increases from $0$ to $1$, and then decays back to $0$.
We reduced (``dampened'') the amplitude of the pseudo-derivative by a factor  $< 1$ (see Suppl.~for details).
This enhances the performance of BPTT for RSNNs that compute during larger time spans,
that require backpropagation through several $1000$ layers of an unrolled feedforward network of spiking neurons.
A similar implementation of BPTT for RSNNs was proposed in \cite{huh2017gradient}.
It is not yet clear which of these two versions of BPTT work best for a given task and a given network.

In order to optimize not only the synaptic weights of a RSNN but also its connectivity matrix, we integrated BPTT with the biologically inspired \cite{kappel2017reward} rewiring method DEEP R \cite{deepr}  (see Suppl.~for details). DEEP R converges theoretically to an optimal network configuration by continuously updating the set of active connections \cite{kappel_network_2015,kappel2017reward,deepr}.

\section{Computational performance of LSNNs}\label{sec:comp_cap}

\begin{figure}
	\includegraphics[width=\textwidth]{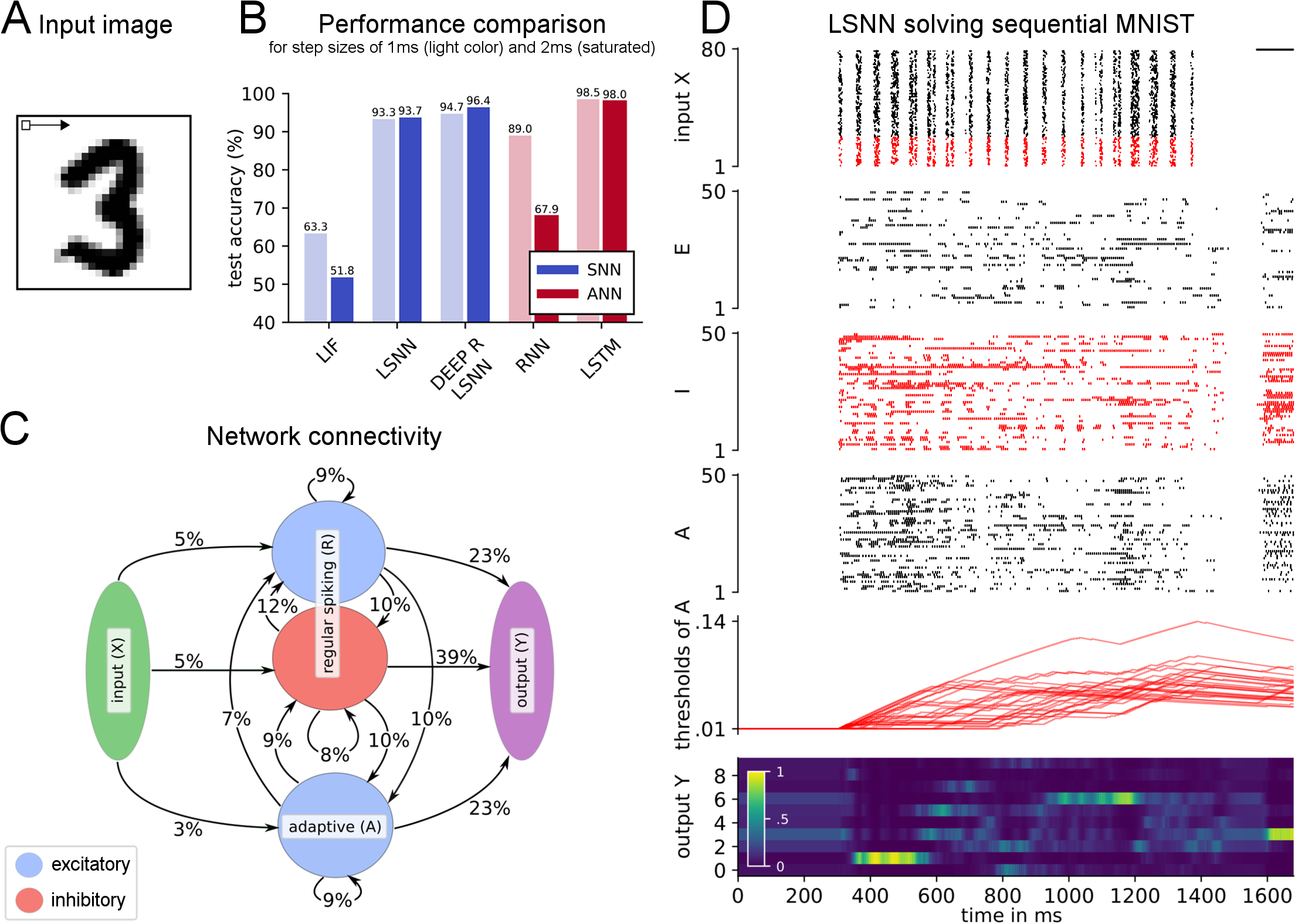}
        \caption{\label{fig:mnist} \textbf{Sequential MNIST.}
          \textbf{A} The task is to classify images of
          handwritten digits when the pixels are shown sequentially pixel by pixel, in a fixed order
          row by row.  \textbf{B} The performance of RSNNs is tested for three different setups: without
          adapting neurons (LIF), a fully connected LSNN, and an LSNN with randomly
          initialized connectivity that was rewired during training (DEEP R LSNN).
          For comparison, the performance of two ANNs, a fully connected RNN and an LSTM network are also shown.
          \mbox{\textbf{C} Connectivity} (in terms of connection probabilities between and within the 3 subpopulations) of the LSNN after applying DEEP R in conjunction with BPTT.
          The input population $X$ consisted of 60 excitatory and 20 inhibitory neurons.
          Percentages on the arrows from $X$ indicate the average connection probabilities from excitatory and inhibitory neurons.
          \textbf{D} Dynamics of the LSNN after training when the input image from A was sequentially presented. From top to
          bottom: spike rasters from input neurons (X), and random subsets of excitatory (E) and inhibitory (I) regularly spiking neurons, and
          adaptive neurons (A), %only $50$ neurons are shown;
          dynamics of the firing thresholds of a random sample of adaptive neurons; activation of softmax readout neurons.}
\end{figure}

\textbf{Sequential MNIST:}
We tested the performance of LSNNs on a standard benchmark task that requires continuous updates of short term memory over a long time span: sequential MNIST \cite{LeJH15, costa2017cortical}.
%The performance of LSNNs is then tested on an artificial benchmark task that was used to compare the performance of competitive artificial neural networks such as LSTMs \cite{le2015simple,costa2017cortical}.
We compare the performance of LSNNs with that of LSTM networks.
The size of the LSNN, in the case of full connectivity, was chosen to match the number of parameters of the LSTM network.
This led to %{\color{red}$120$} 
$120$ regular spiking and $100$ adaptive neurons (with adaptation time constant $\tau_a$ of $700$~ms) in comparison to $128$ LSTM units.
Actually it turned out that the sparsely connected LSNN shown in Fig.~\ref{fig:mnist}C,
which was generated by including DEEP R in BPTT, had only $12\%$ of the synaptic connections but performed better than the fully connected LSNN (see ``DEEP R LSNN'' versus ``LSNN'' in Fig.~\ref{fig:mnist}B).

The task is to classify the handwritten digits of the MNIST dataset when the pixels of each handwritten digit are presented sequentially, one after the other in $784$ steps, see Fig.~\ref{fig:mnist}A.
After each presentation of a handwritten digit, the network is required to output the corresponding class.
The grey values of pixels were given directly to artificial neural networks (ANNs), and encoded by spikes for RSNNs. %(one pixel per $1$ ms).
We considered both the case of step size $1$~ms (requiring $784$~ms for presenting the input image) and $2$~ms (requiring $1568$~ms for each image, the adaptation time constant $\tau_a$ was set to $1400$~ms in this case, see Fig.~\ref{fig:mnist}B.).
The top row of  Fig.~\ref{fig:mnist}D shows a version where the grey value of the currently presented pixel is encoded by population coding through the firing probability of the $80$ input neurons.
Somewhat better performance was achieved when each of the $80$ input neurons is associated with a particular threshold for the grey value, and this input neuron fires whenever the grey value crosses its threshold in the transition from the previous to the current pixel (this input convention is chosen for the SNN %{\color{red}SNN} 
results of Fig.~\ref{fig:mnist}B).
In either case, an additional input neuron becomes active when the presentation of the $784$ pixel values is finished, in order to prompt an output from the network.
The firing of this additional input neuron is shown at the top right of the top panel of Fig.~\ref{fig:mnist}D.
The softmax of 10 linear output neurons $Y$ is trained through BPTT to produce, during this time segment, the label of the sequentially presented handwritten digit.
We refer to the yellow shading around $800$~ms of the output neuron for label 3 in the plot of the dynamics of the output neurons $Y$ in Fig.~\ref{fig:mnist}D. This output was correct.
%Fig.~\ref{fig:mnist}C (bottom) shows that the firing thresholds of adapting neurons undergo complex temporal modifications.

%The output of the network is determined by averaging the readout output over the 56 ms following the presentation of the digit. The network is trained by minimizing the cross entropy error between the softmax of the averaged readout and the label distributions.

A performance comparison is given in Fig.~\ref{fig:mnist}B. LSNNs achieve %{\color{red}$94.7\%$ and $96.4\%$}
$94.7\%$ and $96.4\%$ classification accuracy on the test set when every pixel is presented for $1$ and $2$ms respectively.
An LSTM network achieves $98.5\%$ and $98.0\%$ accuracy on the same task setups.
%The same performance is achieved by an LSTM receiving continuous grey levels instead of spikes. 
The LIF and RNN bars in Fig.~\ref{fig:mnist}B show that this accuracy is out of reach for BPTT applied to spiking or nonspiking neural networks
%with the same number of weights but
without enhanced short term memory capabilities.
We observe that in the sparse architecture discovered by DEEP R, the
connectivity onto the readout neurons Y is denser than in the rest of the network (see Fig.~\ref{fig:mnist}C).
Detailed results are given in the supplement.

%Standard artificial RNN achieve $75\%$, and spiking recurrent networks of regular spiking neurons achieve $61 \%$ in the absence of adaptation.

%Each image presentation with this setup takes 784 ms. To test if the LSNN can also integrate information over a longer time span, we expanded this duration to $3.36$~s by presenting each pixel for $4$~ms instead of $1$. An LSNN trained on this expanded presentation duration achieved in a first experiment $83\%$ classification accuracy (all adaptive neurons had here an adaptation time constant of $3500$~ms).

%\textbf{-- Possible investigation of the dependency with respect to parameters $\beta$, $\gamma$ and rewiring connectivity}

\textbf{Speech recognition (TIMIT):}
We also tested the performance of LSNNs for a real-world speech recognition task, the TIMIT dataset.
A thorough study of the performance of many variations of LSTM networks on TIMIT has recently been carried out in \cite{greff2017lstm}.
We used exactly the same setup %of the TIMIT task 
which was used there (framewise classification) in order to facilitate comparison.
We found that a standard LSNN consisting of $300$ regularly firing ($200$ excitatory and $100$ inhibitory) and $100$ excitatory adapting neurons with an adaptation time constant of $200$~ms, and
with $20\%$ connection probability in the network, achieved a classification error of $33.2\%$.
This error is below the mean error around $40\%$ from $200$ trials with different hyperparameters for the best performing (and most complex)
version of LSTMs according to Fig.~3 of \cite{greff2017lstm}, but above the mean of $29.7\%$ of the $20$ best performing choices of hyperparameters for these LSTMs.
The performance of the LSNN was however somewhat better than the error rates achieved in \cite{greff2017lstm} for a less complex version of LSTMs without forget gates (mean of the best $20$ trials: $34.2\%$).

%\nopagebreak
We could not perform a similarly rigorous search over LSNN architectures and meta-parameters as was carried out in \cite{greff2017lstm} for LSTMs.
But if all adapting neurons are replaced by regularly firing excitatory neurons one gets a substantially higher error rate than the LSNN with adapting neurons: $37\%$. Details are given in the supplement.

\section{LSNNs learn-to-learn from a teacher}\label{sec:L2L}

One likely reason why learning capabilities of RSNN models have remained rather poor is that one usually requires a tabula rasa RSNN model to learn.
In contrast, RSNNs in the brain have been optimized through a host of preceding processes, from evolution to prior learning of related tasks, for their learning performance. 
% Therefore we initiate here an application of learning-to-learn (L2L) to RSNNs.
We emulate a similar training paradigm for RSNNs using the L2L setup.
We explore here only the application of L2L to LSNNs, but L2L can also be applied to RSNNs without adapting neurons \cite{subramoney_etal_2018}.
An application of L2L to LSNNs is tempting, since L2L is most commonly applied in machine learning to their ANN counterparts: LSTM networks
see e.g. \cite{wang2016learning,duan2016rl}.
LSTM networks are especially suited for L2L since they can accommodate two levels of learning and representation of learned insight: Synaptic connections and weights can encode, on a higher level, a learning algorithm and prior knowledge on a large time-scale.
The short-term memory of an LSTM network can accumulate, on a lower level of learning, knowledge during the current learning task.
It has recently been argued \cite{WangETAL:18} that the pre-frontal cortex (PFC) similarly accumulates knowledge during fast reward-based learning in its short-term memory, without using dopamine-gated synaptic plasticity, see the text to Suppl. Fig. 3 in \cite{WangETAL:18}. The experimental results of \cite{perich2018neural} suggest also a prominent role of short-term memory for fast learning in the motor cortex.

\begin{figure}[p]
	\includegraphics[width=\textwidth]{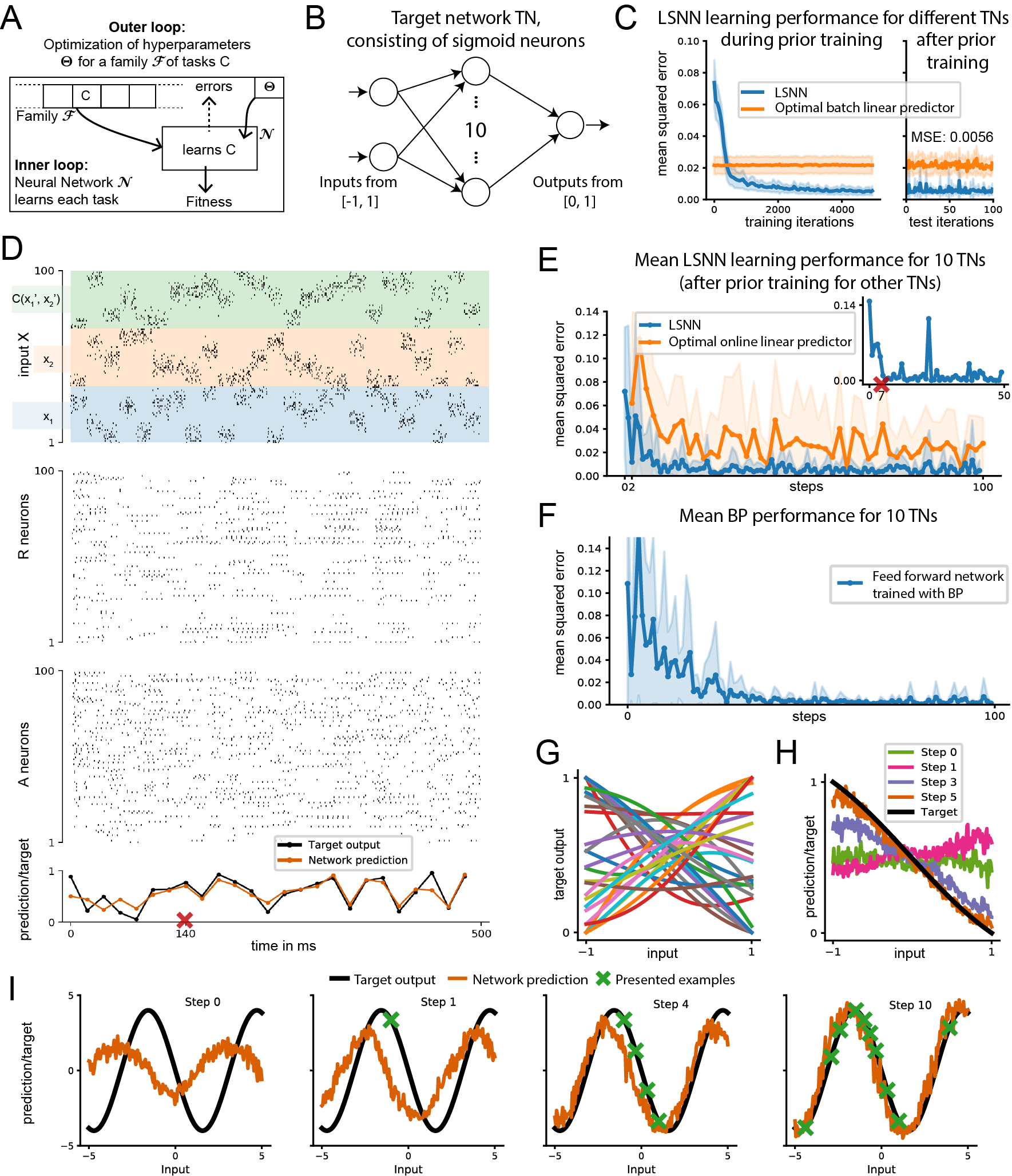}
% 	  \caption{(Caption next page.)}
% \end{figure}
% %\clearpage % insert a page break
% \addtocounter{figure}{-1}
% \begin{figure}
	\caption{
		\label{fig:ltl} \textbf{LSNNs learn to learn from a teacher.} 
        \textbf{A} L2L scheme for an SNN $\mathcal{N}$.
		\textbf{B} Architecture of the two-layer feed-forward target networks (TNs) used to generate nonlinear functions for the LSNN to learn; weights and biases were randomly drawn from [-1,1].
        \textbf{C} Performance of the LSNN in learning a new TN during (left) and after (right) training in the outer loop of L2L. Performance is compared to that of an optimal linear predictor fitted to the batch of all 500 experiments for a TN.
		%The architecture of the LSNN used for the learning to learn task. It received as inputs at each step $x_1, x_2$ and the target from the previous step $C(x_1', x_2')$. Its prediction $\hat{C}(x_1, x_2)$ was output through a linear readout.
		\textbf{D} Network input (top row, only 100 of 300 neurons shown),
		internal spike-based processing with low firing rates in the populations R and A (middle rows), and network output (bottom row) for 25 trials of 20 ms each. 
		%Performance during training and testing. A new function (target network TN) was randomly chosen for each episode. The LSNN is enabled by its weights $\mathbf{W}$ to learn approximations to the input-output behaviour of target networks TN that was significantly better than the best linear approximation to its input-output behaviour.
        \textbf{E} Learning performance of the LSNN for 10 new TNs. Performance for a single TN is shown as insert, a red cross marks step 7 after which output predictions became very good for this TN. The spike raster for this learning process is the one depicted in C. Performance is compared to that of an optimal linear predictor, which, for each example, is fitted to the batch of all preceding examples.
		%The first 100 steps in a sample test episode where the network initially had high error, but after few steps learned to predict the new targets successfully. 
		\textbf{F} Learning performance of BP for the same 10 TNs as in D, working directly on the ANN from A, with a prior for small weights. 
		%Its learning progress is slower compared with that of the LSNN shown in D.
		%The correlation of the output of the LSNN vs target outputs at the start, middle and end of
		%training. 
		%The initial network outputs are random, and as training progresses, the network learns to predict the
		%target better.
		\textbf{G} Sample input/output curves of TNs on a 1D subset of the 2D input space, for different weight and bias values.
		\textbf{H} These curves are all fairly smooth, like the internal models produced by the LSNN while learning a particular TN.
		\textbf{I} Illustration of the prior knowledge acquired by the LSNN through L2L for another family $\mathcal{F}$ (sinus functions). Even adversarially chosen examples (Step 4) do not induce the LSNN to forget its prior.
		%Sample spike raster during testing, of the input neurons (every third neuron, top panel), and the
		%regular (middle panel) and adaptive neuron populations (bottom panel) for the same episode as in (\textbf{D}).
		%The point where the error goes to 0 is marked by a red cross corresponding to the one in (\textbf{D}).
	} 
\end{figure}

The standard setup of L2L involves a large, in fact in general infinitely large, family $\mathcal{F}$ of learning tasks $C$.
Learning is carried out simultaneously in two loops (see Fig.~\ref{fig:ltl}A).
The {\em inner loop} learning involves the learning of a single task $C$ by a neural network $\mathcal{N}$, in our case by an LSNN. %This inner loop learning process is intended to be biologically realistic in our work. 
Some parameters of $\mathcal{N}$ (termed hyper-parameters) are optimized in an {\em outer loop} optimization to support fast learning of a randomly drawn task $C$ from $\mathcal{F}$.
The outer loop training -- implemented here through BPTT -- proceeds on a much larger time scale than the inner loop, integrating performance evaluations from many different tasks $C$ of the family $\mathcal{F}$.  
%insert Einfügung+ - WM 18.5.18
One can interpret this outer loop as a process that mimics the impact of evolutionary and developmental optimization processes, as well as prior learning, on the learning capability of brain networks. 
We use the terms training and optimization interchangeably, but the term training is less descriptive of the longer-term evolutionary processes we mimic.
Like in \cite{hochreiter2001learning,wang2016learning,duan2016rl} we let all synaptic weights of $\mathcal{N}$ belong to the set of hyper-parameters that are optimized through the outer loop.
Hence the network is forced to encode all results from learning the current task $C$ in its internal state, in particular in its firing activity and the thresholds of adapting neurons.
Thus the synaptic weights of the neural network $\mathcal{N}$ are free to encode an efficient \textit{algorithm} for learning arbitrary tasks $C$ from $\mathcal{F}$. %, rather than results of learning a particular task $C$. 

When the brain learns to predict sensory inputs, or state changes that result from an action, this can be formalized as learning from a teacher (i.e., supervised learning).
The teacher is in this case the environment, which provides -- often with some delay -- the target output of a network.
The L2L results of \cite{hochreiter2001learning} show that LSTM networks can learn nonlinear functions from a teacher without modifying their synaptic weights, using their short-term memory instead.
We asked whether this form of learning can also be attained by LSNNs.

\textbf{Task:}
We considered the task of learning complex non-linear functions from a teacher. 
Specifically, we chose as family $\mathcal{F}$ of tasks a class of continuous functions of two real-valued variables $(x_1,x_2)$. %Einfügung11
This class was defined as the family of all functions that can be computed by a 2-layer artificial neural network of sigmoidal neurons with 10 neurons in the hidden layer, and weights and biases from [-1, 1], see Fig.~\ref{fig:ltl}B.
Thus overall, each such target network (TN) from $\mathcal{F}$ was defined through 40 parameters in the range [-1, 1]: 30 weights and 10 biases. 
%insert* WM 18.5.18
We gave the teacher input to the LSNN for learning a particular TN $C$ from $\mathcal{F}$ in a delayed manner as in \cite{hochreiter2001learning}: The target output value was given after $\mathcal{N}$ had provided its guessed output value for the preceding input.

This delay of the feedback is consistent with biologically plausible scenarios. 
Simultaneously, having a delay for the feedback prevents $\mathcal{N}$ from passing on the teacher value as output without first producing a prediction on its own.
% This avoids that $\mathcal{N}$ learns to cheat by simply using the teacher input as its output.
%We are using the same convention in our experiment.
%The LSNN consisted here of 100 regular firing neurons (population R) and 100 adapting neurons (population A) with full connectivity.

\textbf{Implementation:}
We considered a LSNN $\mathcal{N}$ consisting of $180$ regularly firing neurons (population R) and $120$ adapting neurons (population A)
with a spread of adaptation time constants sampled uniformly between $1$ and $1000$~ms and with full connectivity.
Sparse connectivity in conjunction with rewiring did not improve performance in this case.
All neurons in the LSNN received input from a population $X$ of 300 external input neurons.
A linear readout received inputs from all neurons in R and A.
The LSNN received a stream of 3 types of external inputs (see top row of Fig.~\ref{fig:ltl}D): the values of $x_1, x_2$, and of the output $C(x_1', x_2')$ of the TN for the preceding input pair $x_1', x_2'$ (set to 0 at the first trial), all represented through population coding in an external population of 100 spiking neurons.
It produced outputs in the form of weighted spike counts during $20$ ms windows from all neurons in the network (see bottom row of Fig.~\ref{fig:ltl}D), where the weights for this linear readout were trained, like all weights inside the LSNN, in the outer loop, and remained fixed during learning of a particular TN.

% \bigskip
% \label{label}

The training procedure in the outer loop of L2L was as follows: Network training was divided into training episodes.
At the start of each training episode, a new 
%set of input pairs $x_1, x_2 \in [-1, 1]$ and new weights (between $[-1, 1]$) for the 
target network TN was randomly chosen and used to generate target values $C(x_1, x_2) \in [0, 1]$ for randomly chosen input pairs $(x_1,x_2)$.
500 of these input pairs and targets were used as training data, and presented one per step to the LSNN during the episode, 
%Einfügung12 - WM 16.5.18
where each step lasted $20$ ms.
LSNN parameters were updated using BPTT to minimize the mean squared error between the LSNN output and the target in the training set, using gradients computed over batches of $10$ such episodes, which formed one iteration of the outer loop.
In other words, each weight update included gradients calculated on the input/target pairs from $10$ different TNs. 
%which prevented the LSNN from 
This training procedure forced the LSNN to adapt its parameters in a way that supported learning of many different TNs, rather than specializing on predicting the output of single TN.
After training, the weights of the LSNN remained fixed, and it was required to learn the input/output behaviour of TNs from $\mathcal{F}$ that it had never seen before in an online manner by just using its short-term memory and dynamics.
See the suppl. for further details.

\textbf{Results:}
Most of the functions that are computed by TNs from the class $\mathcal{F}$ are nonlinear, as illustrated in  Fig.~\ref{fig:ltl}G for the case of inputs $(x_1, x_2)$ with $x_1  =  x_2$.
Hence learning the input/output behaviour of any such TN with biologically realistic local plasticity mechanisms presents a daunting challenge for a SNN.
Fig.~\ref{fig:ltl}C shows that after a few thousand training iterations in the outer loop, %(mimicking for example evolutionary and developmental processes on a functional level), 
the LSNN achieves low MSE for learning new TNs from the family $\mathcal{F}$, significantly surpassing the performance of an optimal linear approximator (linear regression) that was trained on all 500 pairs of inputs and target outputs, see orange curve in Fig.~\ref{fig:ltl}C,E. 
In view of the fact that each TN is defined by 40 parameters, it comes at some surprise that the resulting network learning algorithm of the LSNN for learning the input/output behaviour of a new TN produces in general a good approximation of the TN after just 5 to 20 trials, where in each trial one randomly drawn labelled example is presented.
One sample of a generic learning process is shown in Fig.~\ref{fig:ltl}D.
Each sequence of examples evokes an internal model that is stored in the short-term memory of the LSNN.
Fig.~\ref{fig:ltl}H shows the fast evolution of internal models of the LSNN for the TN during the first trials (visualized for a 1D subset of the 2D input space).
We make the current internal model of the LSNN visible by probing its prediction $C(x_1,x_2)$ for hypothetical new inputs for evenly spaced points $(x_1, x_2)$ in the domain (without allowing it to modify its short-term memory; all other inputs advance the network state according to the dynamics of the LSNN).
One sees that the internal model of the LSNN is from the beginning a smooth function, of the same type as the ones defined by the TNs in $\mathcal{F}$.
Within a few trials this smooth function approximated the TN quite well.
Hence the LSNN had acquired during the training in the outer loop of L2L a prior for the types of functions that are to be learnt, that was encoded in its synaptic weights.
This prior was in fact quite efficient, since Fig.~\ref{fig:ltl}E and F show that the LSNN was able to learn a TN with substantially fewer trials than a generic learning algorithm for learning the TN directly in an artificial neural network as in Fig. 2A: BP with a prior that favored small weights and biases (see end of Sec.~3 in suppl.).
These results suggest that L2L is able to install some form of prior knowledge about the task in the LSNN.
% The acquired learning algorithm might thus fit one of an acquired reservoir of internal models (for smooth functions, as defined by the TNs in $\mathcal{F}$) to the examples it received. 
We conjectured that the LSNN fits internal models for smooth functions to the examples it received.

We tested this conjecture in a second, much simpler, L2L scenario.
Here the family $\mathcal{F}$ consisted of all sinus functions with arbitrary phase and amplitudes between 0.1 and 5. Fig.~\ref{fig:ltl}I shows that the LSNN also acquired an internal model for sinus functions (made visible analogously as in Fig.~\ref{fig:ltl}H) in this setup from training in the outer loop.  
Even when we selected examples in an adversarial manner, which happened to be in a straight line, this did not disturb the prior knowledge of the LSNN.

% Even adversarially selected labeled examples that formed a straight line (third plot in Fig.~\ref{fig:ltl}H) did not erase this prior knowledge of the LSNN: it still produced a sinus function as prediction, which was fitted to these examples. 

Altogether the network learning that was induced through L2L in the LSNNs is of particular interest from the perspective of the design of learning algorithms, since we are not aware of previously documented methods for installing structural priors 
%for input/output behaviours or internal models that can be learnt by an artificial recurrent neural network, even 
for online learning of a recurrent network of spiking neurons.

\section{LSNNs learn-to-learn from reward}\label{sec:meta-RL}

\begin{figure}
	\includegraphics[width=\textwidth]{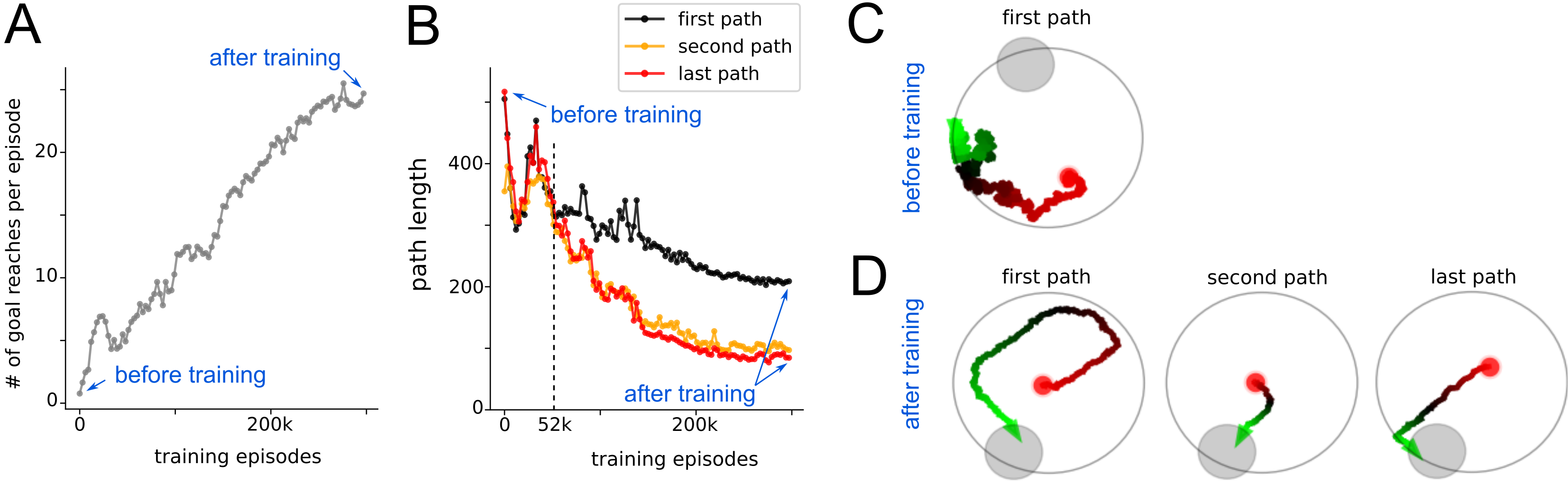}
\caption{\textbf{Meta-RL results for an LSNN.}  \textbf{A, B} Performance improvement during
training in the outer loop. \textbf{C, D} Samples of navigation paths produced by the LSNN before and after this training.
Before training, the agent performs a random walk (\textbf{C}).
In this example it does not find the goal within the limited episode duration.
After training (\textbf{D}), the LSNN had acquired an efficient exploration strategy that uses two pieces of abstract knowledge:
that the goal always lies on the border, and that the goal position is the same throughout an episode. Note that all synaptic weights of the LSNNs remained fixed after training.
} 
\label{fig:maze} 
\end{figure}

%Besides learning-to-learn from a teacher, as discussed in the preceding section, learning-to-
%learn from rewards, or equivalently
We now turn to an application of meta reinforcement learning (meta-RL) to LSNNs. In meta-RL, the LSNN receives rewards instead of teacher inputs.
Meta-RL has led to a
number of remarkable results for LSTM networks, see e.g. \cite{wang2016learning,duan2016rl}. In
addition, \cite{WangETAL:18} demonstrates that meta-RL provides a very interesting
perspective of reward-based learning in the brain. We focused on one of the
more challenging demos of \cite{wang2016learning} and \cite{duan2016rl}, where an agent had
to learn to find a target in a 2D arena, and to navigate subsequently to this target from
random positions in the arena.
This task is related to the well-known biological learning paradigm of the Morris water maze task \cite{morris1984developments,vasilaki2009spike}.
We study here the capability of an agent to discover two pieces of abstract knowledge from the concrete setup of the task: the distribution of goal positions, and the fact that the goal position is constant within each episode.
We asked whether the agent would be able to exploit the pieces of abstract knowledge from learning for many concrete episodes, and use it to navigate more efficiently.

\textbf{Task:} An LSNN-based agent was trained on a family of navigation tasks with continuous state and action spaces in a circular arena. The task is structured as a sequence of episodes, each lasting 2 seconds.
The goal was placed randomly for each episode on the border of the arena.
When the agent reached the goal, it received a reward of 1, and was placed back randomly in the arena.
When the agent hit a wall, it received a negative reward of -0.02 and the velocity vector was truncated to remain inside the arena.
The objective was to maximize the number of goals reached within the episode.
This family $\mathcal{F}$ of tasks is defined by the infinite set of possible goal positions.
For each episode, an optimal agent is expected to explore until it finds the goal position, memorize it and exploits this knowledge until the end of the episode by taking the shortest path to the goal. We trained an LSNN so that the network could control the agent's behaviour in all tasks, without changing its network weights.

%It is remarkable that a RSNN of moderate size can control such highly structured sequential behaviour.

\textbf{Implementation:}
Since LSNNs with just a few hundred neurons are not able to process visual input,
we provided the current position of the agent within the arena through a place-cell like Gaussian population rate encoding of the current position.
The lack of visual input made it already challenging to move along a smooth path, or to stay within a safe distance from the wall. 
The agent received information about positive and negative rewards in the form of spikes from external neurons.
For training in the outer loop, we used BPTT together with DEEP R applied to the surrogate objective of the Proximal Policy Optimization (PPO) algorithm \cite{schulman2017proximal}. 
In this task the LSNN had 400 recurrent units (200 excitatory, 80 inhibitory and 120 adaptive neurons with adaptation time constant $\tau_a$ of $1200$~ms), the network was rewired with a fixed connectivity of $20\%$. The resulting network diagram and spike raster is shown in Suppl.~Fig.~1.
%It received information about rewards in the form of spike inputs, like circuits in the PFC according to \cite{WangETAL:18}.

\textbf{Results:} The network behaviour before, during, and after L2L optimization is shown in Fig.~\ref{fig:maze}.
Fig.~\ref{fig:maze}A shows that a large number of training episodes finally provides significant improvements.
With a close look at Fig.~\ref{fig:maze}B, one sees that before 52k training episodes, the intermediate path planning strategies did not seem to use the discovered goal position to make subsequent paths shorter.
Hence the agents had not yet discovered that the goal position does not change during an episode.
After training for 300k episodes, one sees from the sample paths in Fig.~\ref{fig:maze}D that both pieces of abstract knowledge had been discovered by the agent. The first path in Fig.~\ref{fig:maze}D shows that the agent exploits that the goal is located on the border of the maze. The second and last paths show that the agent knows that the position is fixed throughout an episode.
Altogether this demo shows that meta-RL can be applied to RSNNs, and produces previously not seen capabilities of sparsely firing RSNNs to extract abstract knowledge from experimentation, and to use it in clever ways for controlling behaviour.

\section{Discussion}

We have demonstrated %a new application of Deep Learning in computational neuroscience: 
that deep learning provides a useful new tool for the investigation of networks of spiking neurons:
It allows us to create architectures and learning algorithms for %recurrent networks of spiking neurons (
RSNNs with enhanced computing and learning capabilities.
%, and hence are more similar to RSNNs in the brain from the functional perspective. 
In order to demonstrate this, we adapted BPTT so that it works efficiently for RSNNs, and can be combined with a biologically inspired synaptic rewiring method (DEEP R). We have shown in section \ref{sec:comp_cap} that this method allows us to create sparsely connected RSNNs that approach the performance of LSTM networks on common benchmark tasks for the classification of spatio-temporal patterns (sequential MNIST and TIMIT). This qualitative jump in the computational power of RSNNs was supported by the introduction of adapting neurons into the model. Adapting neurons introduce a spread of longer time constants into RSNNs, as they do in the neocortex according to \cite{allen-featuresearch}. We refer to the resulting variation of the RSNN model as LSNNs, because of the resulting longer short-term memory capability. This form of short-term memory is of particular interest from the perspective of energy efficiency of SNNs, because it stores and transmits stored information through non-firing of neurons: A neuron that holds information in its increased firing threshold tends to fire less often.

%We have shown in sections \ref{sec:L2L} and \ref{sec:meta-RL} that also learning-to-learn (L2L) and meta-reinforcement learning (meta-RL) can be ported in this way into RSNNs. In particular we have shown in Fig.~\ref{fig:ltl} that one arrives in this way at new forms of network learning in RSNNs, that engage their short-term memory, rather than synaptic plasticity, for fast learning. It was recently argued in \cite{WangETAL:18} that this form of fast learning is implemented in the PFC, and optimized through synaptic plasticity on a longer time scale. 
We have shown in Fig.~\ref{fig:ltl} that an application of deep learning (BPTT and DEEP R) in the outer loop of L2L provides a new paradigm for learning of nonlinear input/output mappings by a RSNN. This learning task was thought to require an implementation of BP in the RSNN. We have shown that it requires no BP, not even changes of synaptic weights.
%in the RSNN.
Furthermore we have shown that this new form of network learning enables RSNNs, after suitable training with similar learning tasks in the outer loop of L2L, to learn a new task from the same class substantially faster. The reason is that the prior deep learning has installed abstract knowledge (priors) about common properties of these learning tasks in the RSNN.
%, so that it does not have to learn a new task from scratch. 
To the best of our knowledge, transfer learning capabilities and the use of prior knowledge (see Fig.~\ref{fig:ltl}I) have previously not been demonstrated for SNNs.
Fig~\ref{fig:maze} shows that %meta-RL can be applied very successfully to RSNNs. 
L2L also embraces the capability of RSNNs to learn from rewards (meta-RL).
For example, it enables a RSNN -- without any additional outer control or clock -- to embody an agent that first searches an arena for a goal, and subsequently exploits the learnt knowledge in order to navigate fast from random initial positions to this goal.
Here, for the sake of simplicity, we considered only the more common case when all synaptic weights are determined by the outer loop of L2L. But similar results arise when only some of the synaptic weights are learnt in the outer loop, while other synapses employ local synaptic plasticity rules to learn the current task \cite{subramoney_etal_2018}.

Altogether we expect that the new methods %for the investigation of computing and learning capabilities of RSNNs 
and ideas that we have introduced will advance our %have substantial benefits %for research on 
understanding and reverse engineering of RSNNs in the brain.
%In addition, they enable energy efficient spike-based computer hardware to reach a previously not attainable level of functionality.
%Einfügung 21 WM 18.5.18
For example, the RSNNs that emerged in Fig.~\ref{fig:mnist}-\ref{fig:maze} all compute and learn with a brain-like sparse firing activity, quite different from a SNN that operates with rate-codes. In addition, these RSNNs present new functional uses of short-term memory that go far beyond remembering a preceding input as in \cite{mongillo_synaptic_2008}, and suggest new forms of activity-silent memory \cite{stokes_activity-silentworking_2015}. 

Apart from these implications for computational neuroscience, our finding that RSNNs can acquire powerful computing and learning capabilities with very energy-efficient sparse firing activity provides new application paradigms for spike-based computing hardware through non-firing.

\subsubsection*{Acknowledgments}

This research/project was supported by the HBP Joint Platform, funded from the European Union's Horizon 2020 Framework Programme for Research and Innovation under the Specific Grant Agreement No. 720270 (Human Brain Project SGA1) and under the Specific Grant Agreement No. 785907 (Human Brain Project SGA2). 
We gratefully acknowledge the support of NVIDIA Corporation with the donation of the Quadro P6000 GPU used for this research. 
Research leading to these results has in parts been carried out on the Human Brain Project PCP Pilot Systems at the J{\"u}lich Supercomputing Centre, which received co-funding from the European Union (Grant Agreement No. 604102).
We gratefully acknowledge Sandra Diaz, Alexander Peyser and Wouter Klijn from the Simulation Laboratory Neuroscience of the J{\"u}lich Supercomputing Centre for their support. 
The computational results presented have been achieved in part using the Vienna Scientific Cluster (VSC).

%Research leading to these results has in parts been carried out on the Human Brain Project PCP Pilot Systems at the Juelich Supercomputing Centre, which received co-funding from the European Union (Grant Agreement $\#$604102). We gratefully acknowledge Sandra Diaz, Alexander Payser and Wouter Klijn from the Simulation Laboratory Neuroscience of the J\"ulich Supercomputing Centre for their support. This research was carried out under partial support by the Human Brain Project of the European Union $\#$720270 and $\#$785907. 

\small
\bibliography{library}
\bibliographystyle{unsrt}

\clearpage
\normalsize

\end{document}

\maketitle

\beginsupplement

We provide in this supplement detailed information on the models and simulations of the main text, structured according to the corresponding sections therein. 

\section{LSNN model}\label{sec:sup_model}

\paragraph{Neuron model: }
In continuous time the spike trains $x_i(t)$ and $z_j(t)$ are formalized as sums of Dirac pulses.
Neurons are modeled according to a standard adaptive leaky integrate-and-fire model.
A neuron $j$ spikes as soon at its membrane potential $V_j(t)$ is above its threshold $B_{j}(t)$. At each spike time $t$, the
membrane potential $V_j(t)$ is reset by subtracting the current threshold value $B_{j}(t)$
and the neuron enters a strict refractory period where it cannot spike again.
Importantly at each spike the threshold $B_{j}(t)$ of an adaptive
neuron is increased by a constant $\beta / \tau_{a,j}$.  Then the threshold decays
back to a baseline value $b_{j}^0$. 
Between spikes the membrane voltage $V_j(t)$ and the threshold $B_{j}(t)$ are following the dynamics
\begin{eqnarray}
\tau_m \dot{V}_j(t) & = & - V_j(t) + R_m I_j(t) \label{eq:voltage_continuous} \\
\tau_{a,j} \dot{B}_{j}(t) & = & b_{j}^0 - B_{j}(t),
\end{eqnarray}
where $\tau_m$ is the membrane time constant, $\tau_{a,j}$ is the adaptation time constant and $R_m$ is the membrane resistance. The input current $I_j(t)$ is defined as the weighted sum of spikes from external inputs and other neurons in the network:
\begin{eqnarray}
 I_j(t) & = & \sum_{i} W_{ji}^{in} x_i(t-d_{ji}^{in}) + \sum_{i} W_{ji}^{rec} z_i(t-d_{ji}^{rec}),
 \label{eq:current}
\end{eqnarray}
where $W_{ji}^{in}$ and $W_{ji}^{rec}$ denote respectively the input and the recurrent synaptic weights and  $d_{ji}^{in}$ and  $d_{ji}^{rec}$ the corresponding synaptic delays.
%Note that the weighted sum of recurrent spikes can potentially include incoming spikes from both recurrent populations $R$ and $A$.
All network neurons are connected to a population of readout neurons with weights $W_{kj}^{out}$. 
When network neuron $j$ spikes, the output synaptic strength $W_{kj}^{out}$ is added to the membrane voltage $y_k(t)$ of all readout neurons $k$. $y_k(t)$ also follows the dynamics of a leaky integrator $\tau_m \dot{y}_k(t) = - y_k(t)$.
%The physical units of voltages, currents, delays, Dirac pulses and synaptic weights are respectively Volt, Ampere, Second, Hertz and Coulomb.
%In all experiments except those reported in Fig.~2, the neurons were either excitatory or inhibitory.

%[???? Nothing is said about exc/inhibitory neurons. But it seems like you use them explicitly now.
%Gui: a paragraph is added below.]

\paragraph{Implementation in discrete time:}
Our simulations were performed in discrete time with a time step $\delta t = 1$ ms.
% and time is considered in milliseconds.
In discrete time, the spike trains are modeled as binary sequences $x_i(t), z_j(t) \in \{0,\frac{1}{\delta t}\}$, so that they converge to sums of Dirac pulses in the limit of small time steps.
Neuron $j$ emits a spike at time $t$ if it is currently not in a refractory period, and
its membrane potential $V_j(t)$ is above its threshold $B_{j}(t)$.
During the refractory period following a spike, $z_j(t)$ is fixed to 0.
The dynamics of the threshold is defined by $B_{j}(t) = b_{j}^0 + \beta b_j(t)$ where $\beta$ is a constant which scales the deviation $b_j(t)$ from the baseline $b_{j}^0$.
The neural dynamics in discrete time reads as follows
\begin{eqnarray}
 {V}_j(t+ \delta t) & = & \alpha V_{j}(t) + (1 - \alpha) R_m I_j(t) - B_{j}(t) z_j(t) \delta t \\
b_{j}(t + \delta t) & = & \rho_j b_{j}(t) + (1 - \rho_j) z_j(t),
\end{eqnarray}
where $\alpha = \exp( - \frac{\delta t}{\tau_m} )$ and $\rho_j = \exp( - \frac{\delta t}{\tau_{a,j}} )$.
The term $B_{j}(t) z_j(t)  \delta t$ implements the reset of the membrane voltage after each spike.
The current $I_j(t)$ is the weighted sum of the incoming spikes.
The definition of the input current in equation \eqref{eq:current} holds also for discrete time, with the difference that spike trains now assume values in $\{0,\frac{1}{\delta t}\}$.

\section{Applying BPTT with DEEP R to RSNNs and LSNNs}\label{sec:sup_bptt}
\paragraph{Propagation of gradients in recurrent networks of LIF neurons:} In artificial recurrent neural networks such as LSTMs, gradients can be computed with backpropagation through time (BPTT). 
For BPTT in spiking neural networks, complications arise from the non-differentiability of the output of spiking neurons, and from the fact that gradients need to be propagated either through continuous time or through many time steps if time is discretized. 
Therefore, in \cite{courbariaux_binarized_2016,esser_convolutional_2016} it was proposed to use a pseudo-derivative.

\begin{equation}
\frac{d z_j(t)}{d v_j(t)} := \max\{0, 1 - | v_j(t) | \},
\label{eq:SD}
\end{equation}
where $v_j(t)$ denotes the normalized membrane potential $v_j(t)=\frac{V_j(t)- B_{j}(t)}{B_{j}(t)}$.
This made it possible to train deep feed-forward networks of deterministic binary neurons
\cite{courbariaux_binarized_2016,esser_convolutional_2016}. We observed that this convention tends to be unstable for very deep (unrolled) recurrent networks of spiking neurons. 
To achieve stable performance we dampened the increase of back propagated errors through spikes by using a pseudo-derivative of amplitude $\gamma < 1$ (typically $\gamma=0.3$):
\begin{equation}
\frac{d z_j(t)}{d v_j(t)} := \gamma \max\{0, 1 - | v_j(t) | \}.
\label{eq:SparseSD}
\end{equation}
Note that in adaptive neurons, gradients can propagate through many time steps in the dynamic threshold. This propagation is not affected by the dampening.

\paragraph{Rewiring and weight initialization of excitatory and inhibitory neurons:}
In all experiments except those reported in Fig.~2, the neurons were either excitatory or inhibitory.
When the neuron sign were not constrained, the initial network weights were drawn from a Gaussian distribution $W_{ji} \sim \frac{w_0}{\sqrt{n_{in}}} \NN (0,1)$,
where $n_{in}$ is the number of afferent neurons in the considered weight matrix (i.e., the number of columns of the matrix), $\NN (0,1)$ is the zero-mean unit-variance Gaussian distribution
and $w_0$ is a weightscaling factor chosen to be $w_0=\frac{1 \text{Volt}}{R_m}\delta t$.
With this choice of $w_0$ the resistance $R_m$ becomes obsolete but the vanishing-exploding gradient theory \cite{bengio_learning_1994,sussillo2014random} can be used to avoid tuning by hand the scaling of $W_{ji}$.
In particular the scaling $\frac{1}{\sqrt{n_{in}}}$ used above was sufficient to initialize networks with realistic firing rates and that can be trained efficiently.

When the neuron sign were constrained, all outgoing weights $W_{ji}^{rec}$ or $W_{ji}^{out}$ of a neuron $i$ had the same sign.
In those cases, DEEP R \cite{deepr} was used as it maintains the sign of each synapse during training.
The sign is thus inherited from the initialization of the network weights.
This raises the need of an efficient initialization of weight matrices for given fractions of inhibitory and excitatory neurons.
To do so, a sign $\kappa_i\in \{-1,1\}$ is generated randomly for each neuron $i$ by sampling from a Bernoulli distribution.
The weight matrix entries are then sampled from $W_{ji} \sim \kappa_i |\NN (0,1)|$ and post-processed to avoid exploding gradients.
Firstly, a constant is added to each weight so that the sum of excitatory and inhibitory weights onto each neuron $j$ $(\sum_i W_{ji})$ is zero \cite{rajan_eigenvalue_2006} (if $j$ has no inhibitory or no excitatory incoming connections this step is omitted).
To avoid exploding gradients it is important to scale the weight so that the largest eigenvalue is lower of equal to $1$ \cite{bengio_learning_1994}.
Thus, we divided $W_{ji}$ by the absolute value of its largest eigenvalue.
When the matrix is not square, eigenvalues are ill-defined. Therefore, we first generated a large enough square matrix and selected the required number of rows or columns with uniform probabilities.
The final weight matrix is scaled by $w_0$ for the same reasons as before.

To initialize matrices with a sparse connectivity, dense matrices were generated as described above and multiplied with a binary mask.
The binary mask was generated by sampling uniformly the neuron coordinates that were non-zero at initialization.
DEEP R maintains the initial connectivity level throughout training by dynamically disconnecting synapses and reconnecting others elsewhere.
The $L_1$-norm regularization parameter of DEEP R was set to $0.01$ and the temperature parameter of DEEP R was left at $0$.

\section{Computational performance of LSNNs}

\paragraph{MNIST setup:}
The pixels of an MNIST image were presented sequentially to the LSNN
in $784$ time steps.  
Two input encodings were considered. First, we
used a population coding where the grey scale value (which is in the range $[0, 1]$) of the currently
presented pixel was directly used as  the firing probability of each of the $80$
input neurons in that time step.

In a second type of input encoding -- that is closer to the way how
spiking vision sensors encode their input -- each of the $80$ input
neurons was associated with a particular threshold for the grey value,
and this input neuron fired whenever the grey value of the currently
presented pixel crossed its threshold. Here, we used two input neurons
per threshold, one spiked at threshold crossings from below, and one
at the crossings from above. This input convention was chosen for the
LSNN results of Fig.~1.B.

The output of the network was determined by averaging the readout
output over the $56$ time steps following the presentation of the digit. The
network was trained by minimizing the cross entropy error between the
softmax of the averaged readout and the label distributions.
The best performing models use rewiring
with a global connectivity level of $12\%$ was used during training to
optimize a sparse network connectivity structure (i.e., when randomly
picking two neurons in the network, the probability that they would be
connected is $0.12$).  This implies that only a fraction of the
parameters were finally used as compared to a similarly performing
LSTM network.

Tables \ref{table:1} and \ref{table:2} contain the results and details of training runs
where each time step lasted for $1$ ms and $2$ ms respectively.

\begin{table}[h!]
\centering
\begin{tabular}{r|l|l|l|l|l|l|l}
    Model & \# neurons & conn. & \# params & \# runs & mean & std. & max. \\
    \hline
    LSTM & 128 & 100\% & 67850 & 12 & 79.8\% & 26.6\% & 98.5\% \\
    RNN & 128 & 100\% & 17930 & 10 & 71.3\% & 24.5\% & 89\% \\
    LSNN & 100(A), 120(R) & 12\% & 8185 (full 68210) & 12 & 94.2\% & 0.3\% & 94.7\% \\
    LSNN & 100(A), 200(R) & 12\% & 14041 (full 117010) & 1 & - & - & 95.7\% \\
    LSNN & 350(A), 350(R) & 12\% & 66360 (full 553000) & 1 & - & - & 96.1\% \\
    LSNN & 100(A), 120(R) & 100\% & 68210 & 10 & 92.0\% & 0.7\% & 93.3\% \\
    LIF & 220 & 100\% & 68210 & 10 & 60.9\% & 2.7\% & 63.3\% \\
\end{tabular}
\caption{Results on the sequential MNIST task when each pixel is displayed for $1$ms.
For an LSNN, DEEP R is used to optimize the network under a sparse connectivity constraint, we report the number of parameters including and not including the disconnected synapses.}
\label{table:1}
\end{table}

\begin{table}[h!]
\centering
\begin{tabular}{r|l|l|l|l|l|l|l}
    Model & \# neurons & conn. & \# params & \# runs & mean & std. & max. \\
    \hline
    LSTM & 128 & 100\% & 67850 & 12 & 48.2\% & 39.9\% & 98.0\% \\
    RNN & 128 & 100\% & 17930 & 12 & 30\% & 23.6\% & 67.9\% \\
    LSNN & 100(A), 120(R) & 12\% & 8185 (full 68210) & 12 & 93.8\% & 5.8\% & 96.4\% \\
    LSNN & 350(A), 350(R) & 12\% & 66360 (full 553000) & 1 & - & - & 97.1\% \\
    LSNN & 100(A), 120(R) & 100\% & 68210 & 10 & 90.5\% & 1.4\% & 93.7\% \\
    LIF & 220 & 100\% & 68210 & 11 & 34.6\% & 8.8\% & 51.8\% \\
\end{tabular}
\caption{Results on the sequential MNIST task when each pixel is displayed for $2$ms.}
\label{table:2}
\end{table}

\paragraph{TIMIT setup:}
To investigate if the performance of LSNNs can scale to real world problems, we considered the TIMIT speech recognition task.
We focused on the frame-wise classification where the LSNN has to classify each audio-frame to one of the $61$ phoneme classes.

We followed the convention of Halberstadt \cite{halberstadt} for grouping of training, validation, and testing sets ($3696$, $400$, and $192$ sequences respectively).
The performance was evaluated on the \textit{core test set} for consistency with the literature.
Raw audio is preprocessed into $13$ Mel Frequency Cepstral Coefficients (MFCCs) with frame size $10$ ms and on input window of $25$ ms.
We computed the first and the second order derivatives of MFCCs and combined them, resulting in $39$ input channels.
These $39$ input channels were mapped to $39$ input neurons which unlike in MNIST emit continuous values $x_i(t)$ instead of spikes,
and these values were directly used in equation \ref{eq:current} for the currents of the postsynaptic neurons.

Since we simulated the LSNN network in $1$ ms time steps,
every input frame which represents $10$ ms of the input audio signal was fed to the LSNN network for $10$ consecutive $1$ ms steps.
The softmax output of the LSNN was averaged over every $10$ steps to produce the prediction of the phone in the current input frame.
The LSNN was rewired with global connectivity level of $20\%$.

\paragraph{Parameter values:}
 
%Note that the physical units are coherent between continuous and discrete time.
For adaptive neurons, we used $\beta_j=1.8$, and for regular spiking
neurons we used $\beta_j=0$ (i.e.~$B_{j}$ is constant).  The baseline
threshold voltage was $b_{j}^0=0.01$ and the membrane time constant
$\tau_m=20$ ms. Networks were trained using the default Adam optimizer,
and a learning rate initialized at $0.01$.
The dampening factor for training was $\gamma=0.3$.

For sequential MNIST, all networks were trained for $36000$
iterations with a batch size
of $256$. Learning rate was decayed by a factor $0.8$ every $2500$ iterations.
The adaptive neurons in the LSNN had an adaptation time constant $\tau_a = 700$ ms ($1400$ ms) for $1$ ms ($2$ ms) per pixel setup.
The baseline artificial RNN contained $128$ hidden units with the hyperbolic
tangent activation function. The LIF network was formed by a fully
connected population of $220$ regular spiking neurons.

%In store-recall the test performance is computed over sequences of 32 consecutive store-recall pairs and averaged over 32 random input sequences. During training with BPTT the depth of the unrolled network is adapted to the distribution of store-recall delays. For expected delays of $2$, $4$, and  $6$ s it resulted in 3200, 5200 and 8000 unrolled time steps respectively. All networks were trained for $500$ iterations. We used the Adam optimizer \cite{kingma2014adam} with default parameters and a learning rate of $0.01$, with sequences presented in batches of sizes indicated in Table \ref{tab:delays} and \ref{tab:architecture} (as ``batch size''). To avoid unrealistically high firing rates, the loss function contains a regularization term that minimizes the distance of average firing rate between individual neurons and a target firing rate of $10$ Hz.

For TIMIT, the LSNN network consisted of $300$ regular neurons and
$100$ adaptive neurons which resulted in approximately $400000$ parameters.
Network was trained for $80$ epochs with batches of $32$ sequences.
Adaptation time constant of adaptive neurons was set to $\tau_a = 200$ ms.
Refractory period of the neurons was set to $2$ ms,
the membrane time constant of the output Y neurons to $3$ ms,
and the synaptic delay was randomly picked from $\{1,2\}$ ms.

We note that due to the rewiring of the LSNN using DEEP R \cite{deepr} method,
only a small fraction of the weights had non-zero values ($8185$ in MNIST, $\sim 80000$ in TIMIT).

%Additionally during training the connections were rewired using DEEP R
%\cite{deepr} method and $20\%$ target connectivity.
%This means that only $\approx 80$k parameters are effectively used
%(the other $320$k being fixed to value $0$).

\section{LSNNs learn-to-learn from a teacher}

% Experiment setup
\paragraph{Experimental setup:}~\\

\textit{Function families:} The LSNN was trained to implement a regression algorithm on a family of functions $\FF$.
Two specific families were considered: In the first function family, the functions were defined by feed-forward neural networks with $2$ inputs, $1$ hidden layer consisting of $10$ hidden neurons, and $1$ output, where all the parameters (weights and biases) were chosen uniformly randomly between $[-1, 1]$. 
The inputs were between $[-1, 1]$ and the outputs were scaled to be between $[0, 1]$.
We call these networks Target Networks (TNs).
In the second function family, the targets were defined by sinusoidal functions $y=A \sin(\phi + x)$ over the domain $x\in[-5, 5]$.
The specific function to be learned was defined then by the phase $\phi$ and the amplitude $A$, which were chosen uniformly random between $[0, \pi]$ and $[0.1, 5]$ respectively.

\textit{Input encoding:} Analog values were transformed into spiking trains to serve as inputs to the LSNN as follows: 
For each input component, $100$ input neurons are assigned values $m_1,\dots m_{100}$ evenly distributed between the minimum and maximum possible value of the input.
Each input neuron has a Gaussian response field with a particular mean and standard deviation, where the means are uniformly distributed between the minimum and maximum values to be encoded, and with a constant standard deviation.
More precisely, the firing rate $r_i$ (in Hz) of each input neuron $i$ is given by $r_i = r_{max}\,\exp\left(-\frac{(m_i - z_i)^2}{2\,\sigma^2}\right)$, where $r_{max} = 200$ Hz, $m_i$ is the value assigned to that neuron, $z_i$ is the analog value to be encoded, and $\sigma = \frac{(m_{max} - m_{min})}{1000}$, $m_{min}$ with $m_{max}$ being the minimum and maximum values to be encoded.

\textit{LSNN setup and training schedule:}
The standard LSNN model was used, with $300$ hidden neurons for the TN family of learning tasks, and $100$ for the sinusoidal family. Of these, $40\%$ were adaptive in all simulations.
%\Anand{\st{There were recurrent connections within the regular and adaptive populations, and there were interconnections between regular and adaptive neurons, i.e., }}
We used all-to-all connectivity between all neurons (regular and adaptive).
The output of the LSNN was a linear readout that received as input the mean 
% [???? what does that mean?] Anand: Is it clearer now?
firing rate of each of the neurons per step i.e the number of spikes divided by $20$ for the $20$ ms time window that the step consists of.

The network training proceeded as follows: A new target function was randomly chosen for each \textbf{episode} of training, i.e., the parameters of the target function are chosen uniformly randomly from within the ranges above (depending on whether its a TN or sinusoidal).
Each \textbf{episode} consisted of a sequence of $500$ \textbf{steps}, each lasting for $20$ ms.
In each step, one training example from the current function to be learned was presented to the LSNN.
In such a step, the inputs to the LSNN consisted of a randomly chosen vector $\bx$ with its dimensionality $d$ and range determined by the target function being used ($d=2$ for TNs, $d=1$ for sinusoidal target function).
In addition, at each step, the LSNN also got the target value $C(\bx')$ from the previous step, i.e., the value of the target calculated using the target function for the inputs given at the previous step (in the first step, $C(\bx')$ is set to $\bnull$).  

All the weights of the LSNN were updated using our variant of BPTT, once per \textbf{iteration}, where an \textbf{iteration} consists of a batch of $10$ \textbf{episodes}, and the weight updates are accumulated across episodes in an iteration.
The Adam \cite{kingma2014adam} variant of BP was used with standard parameters and a learning rate of $0.001$.  
The loss function for training was the mean squared error (MSE) of the LSNN predictions over an iteration (i.e. over all the steps in an episode, and over the entire batch of episodes in an iteration).
In addition, a regularization term was used to maintain a firing rate of $20$ Hz.
Specifically, the regularization term $R$ is defined as the mean squared difference between the average neuron firing rate in the LSNN and a target of $20$ Hz. The total loss $L$ was then given by $L = MSE + 30\,R$.  
In this way, we induce the LSNN to use sparse firing.
We trained the LSNN for $5000$ iterations in all cases.

\paragraph{Parameter values:}

The LSNN parameters were as follows: $5$ ms neuronal refractory period, delays spread uniformly between $0-5$ ms, adaptation time constants of the adaptive neurons spread uniformly between $1-1000$ ms, $\beta=1.6$ for adaptive neurons ($0$ for regular neurons), membrane time constant $\tau=20$ ms, $0.03$ mV baseline threshold voltage. 
The dampening factor for training was $\gamma=0.4$.

\paragraph{Analysis and comparison:}
% The optimal linear baseline in Fig.~2(B,D,E) was calculated using ridge regression for each episode using the mean trace  per step of the spiking encoding of $\bx$ as input to a linear regressor. 
The linear baseline was calculated using linear regression with L2 regularization with a regularization factor of $100$ (determined using grid search), using the mean spiking trace of all the neurons.
The mean spiking trace was calculated as follows: First the neuron traces were calculated using an exponential kernel with $20$ ms width and a time constant of $20$ ms.
Then, for every step, the mean value of this trace was calculated to obtain the mean spiking trace.
In Fig.~2C, for each episode consisting of $500$ steps, the mean spiking trace from a random subset of $450$ steps was used to train the linear regressor, and the mean spiking trace from remaining $50$ steps was used to calculate the test error. The reported baseline is the mean of the test error over one batch of $10$ episodes with error bars of one standard deviation.
In Fig.~2E, for each episode, after every step $k$, the mean spiking traces from the first $k-1$ steps were used to train the linear regressor, and the test error was calculated using the mean spiking trace for the $k$th step. The reported baseline is a mean of the test error over one batch of $10$ episodes with error bars of one standard deviation.

For the case where neural networks defined the function family, the total test MSE was $0.0056 \pm 0.0039$ (linear baseline MSE was $0.0217 \pm 0.0046$). 
For the sinusoidal function family, the total test MSE was $0.3134 \pm 0.2293$ (linear baseline MSE was $1.4592 \pm 1.2958$).

\textit{Comparison with backprop:}
The comparison was done for the case where the LSNN is trained on the function family defined by target networks.
A feed-forward (FF) network with $10$ hidden neurons and $1$ output was constructed.
% The same exact spiking input given to the LSNN was used to train this FF network. 
The input to this FF network were the analog values that were used to generate the spiking input and targets for the LSNN.
Therefore the FF had $2$ inputs, one for each of $x_1$ and $x_2$.
The error reported in Fig~2F is the mean training error over $10$ batches with error bars of one standard deviation.
% Therefore the FF had $200$ inputs -- $100$ spiking neuron inputs for each of $x_1$, $x_2$ and the previous step target $C(x_1', x_2')$.
% The $20$ ms of spiking input for each input and the corresponding analog target was used as a batch (of size $20$) for one iteration of training for the FF network.
% [???? Spiking input into an artificial neural network ? Anand: Yes, to make the comparison fair in some sense since there is a loss of precision associated with the spiking encoding]

The FF network was initialized with Xavier normal initialization \cite{GlorotBengio:10} (which had the best performance, compared to Xavier uniform and plain uniform between $[-1, 1]$).
Adam \cite{kingma2014adam} with AMSGrad \cite{reddi2018convergence} was used with parameters $\eta = 10^{-1}, \beta_1 = 0.7, \beta_2 = 0.9, C = 10^{-5}$. These were the optimal parameters as determined by a grid search.
Together with the Xavier normal initialization and the weight regularization parameter $C$, the training of the FF favoured small weights and biases.

\section{LSNNs learn-to-learn from reward}

\begin{figure}
\includegraphics[width=\textwidth]{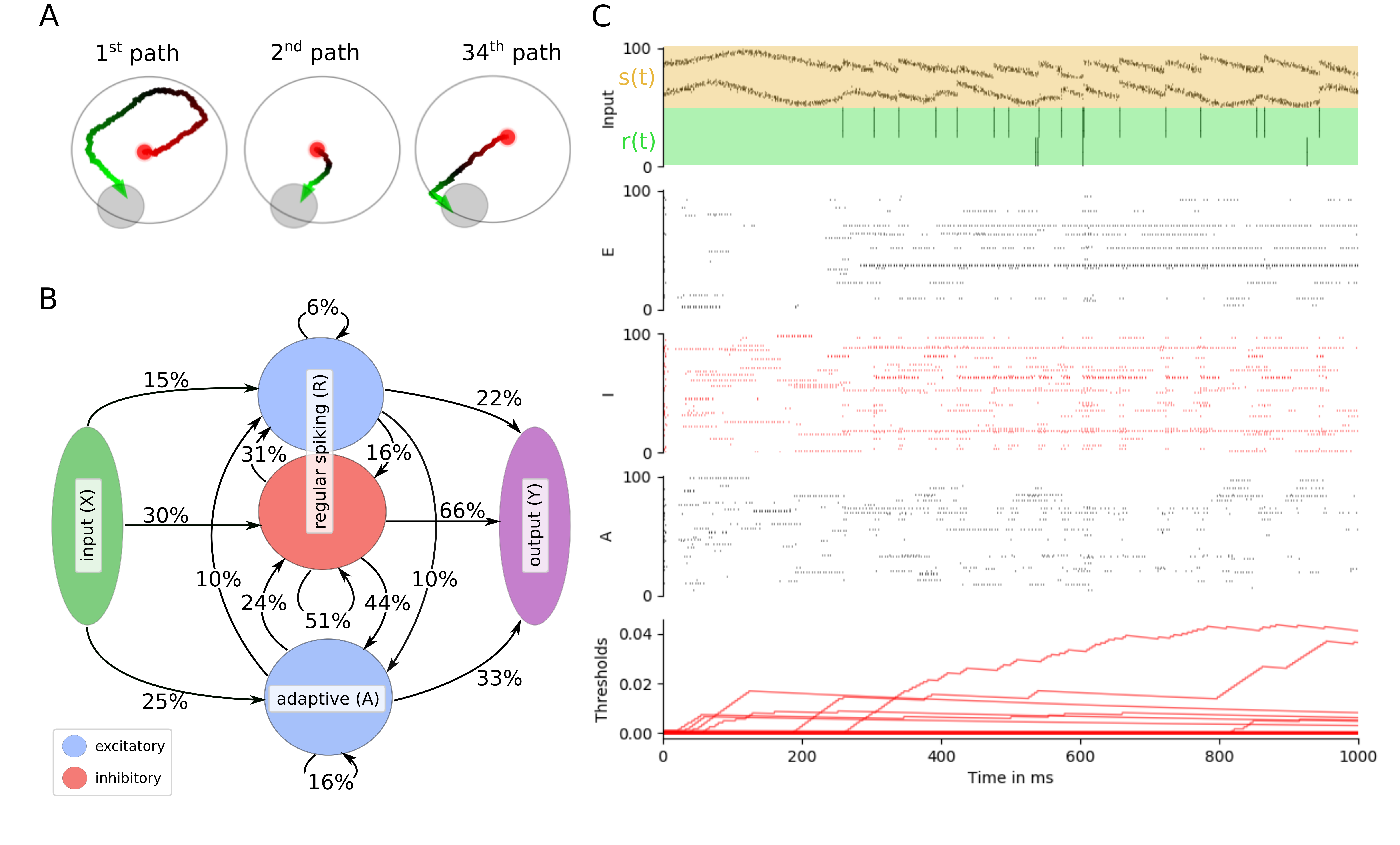}
\caption{\textbf{Meta-RL results for an LSNN.} \textbf{A} Samples of paths after training. \textbf{B} Connectivity between sub-populations of the network after training.
The global connectivity in the network was constrained to $20\%$. \textbf{C} The network dynamics that produced the behavior shown in A. Raster plots and thresholds are displayed as in Fig.~1.D, only 1 second and 100 neurons are shown in each raster plots.}
\label{fig:raster_continuous_maze}
\end{figure}

\paragraph{Experimental setup:}~\\

\textit{Task family: } 
An LSNN-based agent was trained on a family of navigation tasks in a two dimensional circular arena.
%In each episode, the goal was placed at a random position uniformly on the circular border of the arena. The objective was to maximize the number of goal reaches within a given duration of 2 seconds.
%An optimal agent is expected for each new episode to explore until it finds the goal position, memorize it and exploit this knowledge until the end of the episode.
%We train an LSNN so that after training the network can control the agent behaviour in all tasks without changing the network weights.
For all tasks, the arena is a circle with radius $1$ and goals are smaller circles of radius $0.3$ with centres uniformly distributed on the circle of radius $0.85$.
At the beginning of an episode and after the agent reaches a goal, the agent's position is set randomly with uniform probability within the arena.
At every timestep, the agent chooses an action by generating a small velocity vector of Euclidean norm smaller or equal to $a_{scale}=0.02$.
When the agent reaches the goal, it receives a reward of $1$.
If the agent attempts to move outside the arena, the new position is given by the intersection of the velocity vector with the border and the agent receives a negative reward of $-0.02$.

\newcommand{\bth}{\theta}
\newcommand{\LL}{{\mathcal{L}}}

\textit{Input encoding:} Information of the current environmental state
$s(t)$ and the reward $r(t)$ were provided to the LSNN at each time
step $t$ as follows: The state $s(t)$ is given by the $x$ and $y$ coordinate 
of the agent's position (see top of Fig.~\ref{fig:raster_continuous_maze}C).  Each position
coordinate $\xi (t) \in [-1,1]$ is encoded by $40$ neurons which
spike according to a Gaussian population rate code defined as follows:
a preferred coordinate value $\xi_i$, is assigned to each of the $40$ neurons, where $\xi_i$'s are evenly spaced between $-1$ and $1$. 
The firing rate of neuron $i$ is then
given by $r_{max} \exp(- 100 (\xi_i - \xi)^2)$ where $r_{max}$ is
$500$ Hz. The instantaneous
reward $r(t)$ is encoded by two groups of $40$ neurons (see green row at the top of Fig.~\ref{fig:raster_continuous_maze}C). All neuron in the first group spike in synchrony each
time a reward of $1$ is received (i.e., the goal was reached), and the
second group spikes when a reward of $-0.02$ is received (i.e., the agent
moved into a wall).  

\textit{Output decoding:} The output of the LSNN is provided by five
readout neurons.  Their membrane
potentials $y_i(t)$ define the outputs of the LSNN. 
The action vector $\mathbf{a}(t)=(a_x(t),a_y(t))^T$ is sampled from the distribution $\pi_{\bth}$ which
depends on the network parameters $\bth$ through the readouts $y_i(t)$ as follows:
The coordinate $a_x(t)$ ($a_y(t)$) is sampled from a Gaussian distribution with mean $\mu_x=\operatorname{tanh}(y_1(t))$ ($\mu_y=\operatorname{tanh}(y_2(t))$) and variance $\phi_x=\sigma(y_3(t))$ ($\phi_y=\sigma(y_4(t))$).
%, where $\mu_0=\operatorname{tanh}(y_1(t))$, $\mu_1=\operatorname{tanh}(y_2(t))$, $\phi_0=\sigma(y_3(t))$ and $\phi_1=\sigma(y_4(t))$.
The velocity vector that updates the agent's position is then defined as $a_{scale}\, \mathbf{a}(t)$.
If this velocity has a norm larger than $a_{scale}$, it is clipped to a norm of $a_{scale}$.

The last readout output $y_5(t)$ is used to predict the value function
$V_{\bth}(t)$. It estimates the expected discounted sum of future
rewards $R(t)=\sum_{t'>t}\eta^{t'-t}r(t')$, where $\eta=0.99$ is the discount factor and $r(t')$ denotes the reward at time $t'$.  To enable the network to
learn complex forms of exploration we introduced current noise in the
neuron model in this task.  At each time step, we added a small
Gaussian noise with mean $0$ and standard deviation $\frac{1}{R_{m}} \nu_j$ to
the current $I_j$ into neuron $j$.  Here, $\nu_j$ is a network
parameter initialized at $0.03$ and optimized by BPTT alongside the network weights. %[???? what does that mean? is $nu_j$ optimized by backprop?]

\paragraph{Network training:}
To train the network we used the Proximal Policy Optimization algorithm (PPO) \cite{schulman2017proximal}.
For each training iteration, $K$ full episodes of $T$ timesteps were generated with fixed parameters $\bth_{old}$ (here $K=10$ and $T=2000$).
We write the clipped surrogate objective of PPO as $O^{PPO}(\bth_{old},\bth,t,k)$ (this is defined under the notation $L^{CLIP}$ in \cite{schulman2017proximal}).
The loss with respect to $\bth$ is then defined as follows: 
\begin{eqnarray}
\LL(\bth) & = & - \frac{1}{K T} \sum_{k<K} \sum_{t<T} O^{PPO}(\bth_{old},\bth,t,k) + \mu_{v} \left( R(t,k) - V_{\bth}(t,k) \right)^2 \\
  &  &-\mu_e H(\pi_{\bth}(k,t)) + \mu_{firing} \frac{1}{n} \sum_{j} || \frac{1}{K T} \sum_{k,t} z_j(t,k) - f^{0}||^2,
\end{eqnarray}
where $H(\pi_{\bth})$ is the entropy of the distribution $\pi_{\bth}$, $f^{0}$ is a target firing rate of $10$ Hz, and $\mu_{v}$, $\mu_e$, $\mu_{firing}$ are regularization hyper-parameters.
Importantly probability distributions used in the definition of the loss $\LL$ (i.e. the trajectories) are conditioned on the current noises, so that for the same noise and infinitely small parameter change from $\bth_{old}$ to $\bth$ the trajectories and the spike trains are the same.
At each iteration this loss function $\LL$ is then minimized with one step of the ADAM optimizer.

\paragraph{Parameter values:}

In this task the LSNN had 400 hidden units (200 excitatory neurons, 80 inhibitory neurons and 120 adaptive neurons with adaptation time constants $\tau_a = 1200$~ms) and the network was rewired with a fixed global connectivity of $20\%$ \cite{deepr}.
The membrane time constants were similarly sampled between $15$ and $30$ ms.
The adaptation amplitude $\beta$ was set to $1.7$.
The refractory period was set to $3$ ms and delays were sampled uniformly between $1$ and $10$~ms.
The regularization parameters $\mu_v$, $\mu_e$ and $\mu_{firing}$ were respectively $1$, $0.001$, and $100$.
The parameter $\epsilon$ of the PPO algorithm was set to $0.2$.
The learning rate was initialized to $0.01$ and decayed by a factor $0.5$ every $5000$ iterations.
We used the default parameters for ADAM, except for the parameter $\epsilon$ which we set to $10^{-5}$.

\small
\bibliography{library}
\bibliographystyle{unsrt}

\clearpage
\normalsize